\documentclass{article}
\usepackage{ijcai26}

\usepackage{times}
\usepackage{soul}
\usepackage{url}
\usepackage[hidelinks]{hyperref}
\usepackage[utf8]{inputenc}
\usepackage[small]{caption}
\usepackage{graphicx}
\usepackage{amsmath}
\usepackage{amsthm}
\usepackage{booktabs}
\usepackage{algorithm}
\usepackage{algorithmic}
\usepackage[switch]{lineno}

\usepackage{natbib} 
 
\usepackage[T1]{fontenc}    
\usepackage{hyperref}       
\usepackage{booktabs}       
\usepackage{amsfonts}       
\usepackage{nicefrac}       
\usepackage{microtype}      
\usepackage{xcolor}         
 
\usepackage{amssymb}
\usepackage{mathtools}  

\title{
Resilient Load Forecasting under Climate Change: Adaptive Conditional Neural Processes for Few-Shot Extreme Load Forecasting
}

\author{
Chenxi Hu
\and
Yue Ma\and
Yifan Wu\And 
Yunhe Hou\\ 
}

\begin{document}

\maketitle  

\begin{abstract}
Extreme weather can substantially change electricity consumption behavior, causing load curves to exhibit sharp spikes and pronounced volatility. If forecasts are inaccurate during those periods, power systems are more likely to face supply shortfalls or localized overloads, forcing emergency actions such as load shedding and increasing the risk of service disruptions and public-safety impacts. This problem is inherently difficult because extreme events can trigger abrupt regime shifts in load patterns, while relevant extreme samples are rare and irregular, making reliable learning and calibration challenging. We propose AdaCNP, a probabilistic forecasting model for data-scarce condition. AdaCNP learns similarity in a shared embedding space. For each target data, it evaluates how relevant each historical context segment is to the current condition and reweights the context information accordingly. This design highlights the most informative historical evidence even when extreme samples are rare. It enables few-shot adaptation to previously unseen extreme patterns. AdaCNP also produces predictive distributions for risk-aware decision-making without expensive fine-tuning on the target domain.
We evaluate AdaCNP on real-world power-system load data and compare it against a range of representative baselines. The results show that AdaCNP is more robust during extreme periods, reducing the mean squared error by 22\% relative to the strongest baseline while achieving the lowest negative log-likelihood, indicating more reliable probabilistic outputs. These findings suggest that AdaCNP can effectively mitigate the combined impact of abrupt distribution shifts and scarce extreme samples, providing a more trustworthy forecasting for resilient power system operation under extreme events.

\end{abstract}

\section{Introduction}

\begin{figure}[t]
    \centering
    \includegraphics[scale=0.8]{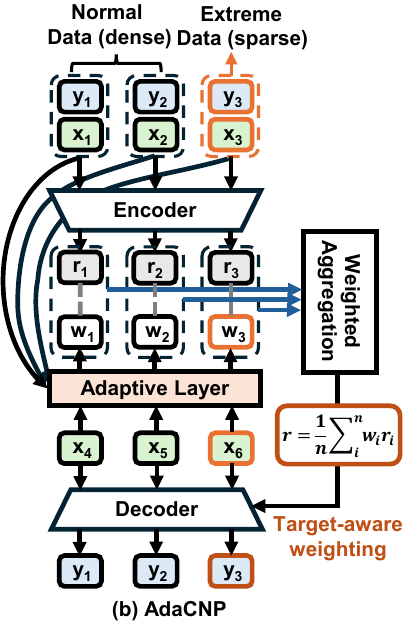}
    \caption{(a) Standard CNP with uniform context aggregation; (b) Our proposed AdaCNP with target-aware adaptive weighting. The adaptive layer assigns relevance scores to context points based on similarity to target extreme events, amplifying critical signals while suppressing noise from irrelevant normal patterns.}  
    \label{fig1}  
\end{figure}

Extreme weather events can substantially influence electricity demand, producing abnormal peaks and highly volatile fluctuations in load curves. When forecasts become inaccurate under such conditions, the power system is more likely to face supply shortfalls or equipment overloads, which increases the risk of curtailment and even outages \citep{haben2023core}. Reliable forecasting during extreme events therefore serves as a critical input to reserve planning, risk assessment, and emergency operations, rather than a routine accuracy improvement for typical conditions.

Two factors make extreme load forecasting particularly challenging. First, extreme events are often accompanied by abrupt distribution shifts, so the input--output relationship learned from normal periods may no longer hold in the extreme regime. Second, extreme samples are inherently scarce and irregular in time, limiting what can be learned or calibrated directly from data \citep{laptev2017time}. These issues are frequently addressed with online or continual adaptation mechanisms \citep{gama2014survey,fekri2021deep,von2020online,bayram2023lstm}, but in extreme periods the available samples can be too limited and too delayed to support reliable updates. Moreover, power system decision making benefits not only from point predictions but also from calibrated uncertainty estimates, which are central to probabilistic forecasting and risk aware operation \citep{liu2015probabilistic,haben2016hybrid,wan2021adaptive,yang2019bayesian}.
Under dataset shift, uncertainty estimates from standard models can become unreliable, which is particularly problematic during extreme events where decisions depend on trustworthy probabilistic information. \citep{ovadia2019can}

Motivated by these requirements, we adopt a conditional generative perspective for probabilistic forecasting. Neural Processes model a distribution over functions with fast amortized inference, combining the flexibility of deep learning with Gaussian process style uncertainty estimation \citep{garnelo2018neural}. Conditional Neural Processes (CNPs) provide a deterministic and computationally efficient variant that predicts a posterior distribution for target inputs given a set of context observations \citep{garnelo2018conditional}. This efficiency and uncertainty awareness make the framework attractive for data limited forecasting tasks.

However, standard CNPs aggregate context points uniformly, implicitly treating all historical contexts as equally informative. Under covariate shift or data sparsity, only a subset of contexts is informative for a given target, and naive averaging can dilute useful signal and degrade posterior accuracy \citep{garnelo2018conditional}. This limitation becomes more pronounced in extreme event forecasting, where normal patterns dominate the training corpus while truly relevant extreme contexts are rare.

A natural remedy is to learn target dependent context weighting, and attentive variants of Neural Processes use multi head attention to provide flexible context aggregation \citep{kim2019attentive}. While attention can increase expressiveness, it typically introduces more complex interactions and additional degrees of freedom. In data scarce and strongly shifted regimes, this added flexibility can raise estimation variance and increase computational overhead, which is undesirable when stable and efficient inference is needed at critical times. These considerations motivate a simpler mechanism that retains the core benefits of Neural Processes while improving target specific adaptation.

In this work, we propose AdaCNP, an adaptive probabilistic forecasting framework designed for data scarce, high variability scenarios. Instead of relying on extensive target domain fine tuning, AdaCNP performs few shot adaptation by learning similarity in a unified embedding space and dynamically recalibrating the importance of historical context points for each target query. Concretely, AdaCNP maps both context pairs and target inputs to latent representations, computes target conditioned weights from their similarity, and forms a target specific aggregated representation instead of uniform averaging. The weighting retains permutation invariance over context sets and remains compatible with the standard CNP decoding pipeline \citep{garnelo2018conditional}. We further introduce a temperature controlled mechanism that smoothly interpolates between near uniform aggregation and highly selective reweighting, enabling stable adaptation across different levels of distribution shift.

Our contributions are summarized as follows:
\begin{itemize}
    \item We propose AdaCNP for probabilistic load forecasting under extreme weather conditions. Built upon the Conditional Neural Process framework, AdaCNP outputs full predictive distributions rather than point forecasts, making it directly useful for power-system applications such as reserve planning, risk assessment, and emergency operations.

    \item We design a target-conditioned context selection and permutation-invariant aggregation module and integrate it into the encoder-decoder pipeline of AdaCNPs. The module assigns weights to historical context points based on input similarity, so that rare contexts relevant to extreme operating conditions are emphasized even when normal patterns dominate the data.  

    \item We conduct a systematic evaluation of AdaCNP on real-world power system load data, with a focus on peak demand and highly volatile periods. In addition to standard test splits, we construct an extreme-period, few-context evaluation setting, where the model must produce probabilistic forecasts under limited information and abrupt distribution shifts. The results show that AdaCNP yields more reliable predictive distributions and achieves better performance at critical time periods, providing a more trustworthy forecasting input for reserve planning, risk assessment, and emergency operation.
\end{itemize}

The remainder of this paper is organized as follows. Section~2 reviews related work and background. Section~3 presents AdaCNP and its adaptive aggregation mechanism. Section~4 reports experimental results. Section~5 concludes and discusses future directions.

\vspace{-0.5em}

\section{Related Work}

\subsection{Load Forecasting and Probabilistic Forecasting}
Accurate load forecasting is a foundational capability for reliable and economical power system operation. Classical learning-based approaches such as neural networks and SVMs were among the early attempts to capture nonlinear consumption patterns beyond purely statistical models \citep{hippert2001neural,chen2004load}. More recently, sequence models including LSTMs and transformer-based architectures have become standard tools for short-term load forecasting due to their ability to model temporal dependencies and complex exogenous effects \citep{kong2017short,wang2022transformer}. Despite strong performance on typical conditions, many of these models are trained under implicit stationarity assumptions and can degrade when the operating regime changes abruptly, which is common during extreme events \citep{liu2023sadi}.

For risk-aware grid operation, point forecasts alone are often insufficient. Probabilistic load forecasting aims to quantify predictive uncertainty, which is critical for reserve scheduling, contingency assessment, and robust decision-making \citep{hong2016probabilistic, haben2023core}. Traditional probabilistic techniques include quantile regression, kernel density estimation, and ensemble-based approaches \citep{liu2015probabilistic,haben2016hybrid,wan2021adaptive}. Recent neural probabilistic methods extend this line with Bayesian neural networks and flow-based density models to better capture complex predictive distributions \citep{yang2019bayesian,wen2022continuous}. These advances motivate models that can produce calibrated uncertainty while remaining reliable under rare and rapidly changing conditions.

\subsection{Extreme Events Forecasting with Distribution Shift}
A large body of work studies distribution shift in time series through concept drift and online adaptation. Concept drift methods emphasize detecting changes over time and updating models incrementally \citep{gama2014survey}. In load forecasting, online and continual learning strategies have been explored to mitigate performance decay without full retraining, including adaptive recurrent models and online ensembles \citep{fekri2021deep,von2020online,bayram2023lstm}. Meta-learning has also been investigated to enable faster adaptation across changing environments \citep{zhu2024learning}. While effective under gradual shifts or when sufficient recent data are available, these approaches can be unreliable during extreme events, where informative samples are scarce, labels may be delayed, and rapid updates can be unstable.

Parallel efforts in rare-event forecasting leverage extreme value theory and imbalance-aware learning to model tail behavior \citep{haan2006extreme,li2019use,gu2021imbalance}. Data augmentation and transfer learning have also been used to improve robustness to rare conditions \citep{bhatia2021exgan,hu2022black}. However, many such methods require substantial retraining, careful domain expertise, or access to labeled target data, which is often impractical in time-critical extreme-event settings. More broadly, OOD generalization and domain adaptation develop principles for handling distribution changes, but commonly rely on multiple domains or labeled target data for tuning and validation \citep{arjovsky2019invariant,liu2021towards,WANG2018135,zhuang2020comprehensive}. These limitations motivate approaches that can adapt to an unseen extreme regime with minimal target information and without expensive updates, while still producing uncertainty estimates suitable for decision-making.

\subsection{Neural Processes}
Neural Processes (NPs) provide a conditional generative framework that combines the representation learning strengths of neural networks with function-space uncertainty modeling in the spirit of Gaussian processes \citep{garnelo2018neural}. Conditional Neural Processes (CNPs) are a tractable deterministic variant that amortizes inference by encoding a set of context observations and decoding a predictive distribution at target inputs \citep{garnelo2018conditional}. This perspective is attractive for data-limited settings because it naturally supports conditional probabilistic prediction with fast inference.

A key limitation of standard CNPs is their uniform context aggregation, which implicitly treats all context points as equally informative. Under distribution shift, only a small subset of historical contexts may be relevant to a specific target query, and uniform averaging can dilute informative signals and degrade conditional inference \citep{garnelo2018conditional}. Attentive Neural Processes address this by learning target-dependent aggregation with multi-head attention \citep{kim2019attentive}. While attention increases expressiveness, it also introduces heavier interactions and additional degrees of freedom, which can be undesirable in rare-event regimes where data are scarce and stable, efficient inference is required.

Our work builds on this line by developing a lightweight, similarity-driven, target-conditioned reweighting mechanism for Neural Processes that is designed specifically for extreme-event forecasting under severe shift. The goal is to retain the efficiency and uncertainty-aware inference of the NP framework while improving target-specific conditioning in the presence of rare, high-impact regimes.

\section{AdaCNP: An Adaptive Conditional Neural Process for Forecasting under Extreme Scenarios} 

In this section, we introduce \textit{AdaCNP} model, which enhances CNPs with adaptive context weighting for robust load forecasting under extreme scenarios. Figure~\ref{fig1} illustrates the architecture, which consists of three key components: (1) a shared encoder for context/target embeddings, (2) an adaptive layer, and (3) a weighted aggregation decoder. The model dynamically computes relevance weights between target queries and historical contexts, enabling focused attention on critical patterns during extreme events.

\subsection{Problem Formulation}
Let $(\mathbf{x}, \mathbf{y}) \sim \mathcal{P}$ denote an input--output pair drawn from an unknown joint distribution $\mathcal{P}$, where $\mathbf{x}\in\mathcal{X}\subseteq \mathbb{R}^{d_x}$ and $\mathbf{y}\in\mathcal{Y}\subseteq \mathbb{R}^{d_y}$.
In load forecasting, $\mathbf{x}$ collects domain features such as lagged load patterns, weather variables, and seasonal/calendar indicators, and $\mathbf{y}$ denotes the target load quantity to be predicted.

We consider supervised regression-style forecasting in which the predictive target $\mathbf{y}$ is modeled conditional on its associated feature vector $\mathbf{x}$.
Given a set of historical observations, our goal is to infer the conditional predictive distribution of target outputs for new inputs.
Throughout, we use the term \emph{context} to denote conditioning observations and the term \emph{target} to denote query points to be predicted.

\paragraph{Historical and target sets.}
We assume a historical dataset $\mathcal{D}_{\mathcal{H}}=\{(\mathbf{x}_m,\mathbf{y}_m)\}_{m=1}^{N_H}$ available for training and context construction, and a separate target dataset $\mathcal{D}_{\mathcal{T}}=\{(\mathbf{x}_m,\mathbf{y}_m)\}_{m=1}^{N_T}$ used for evaluation.
We focus on targets corresponding to extreme scenarios, where the conditional relationship $p(\mathbf{y}\mid \mathbf{x})$ may change abruptly relative to typical regimes and informative samples are sparse.

In practice, the boundary between normal and extreme regimes is unknown.
We therefore use a data-driven event detector to assign each example an event label (normal or extreme), and denote the corresponding partitions as
\begin{equation}
\mathcal{D}_{\mathcal{H}}=\mathcal{D}_{\mathcal{H}}^{e}\cup \mathcal{D}_{\mathcal{H}}^{n},\qquad
\mathcal{D}_{\mathcal{T}}=\mathcal{D}_{\mathcal{T}}^{e}\cup \mathcal{D}_{\mathcal{T}}^{n},
\end{equation}
where superscripts $e$ and $n$ indicate extreme and normal subsets, respectively.
The specific detector (e.g., a 3-$\sigma$ rule applied to a distance-based outlier score) is described in the experimental section.

\paragraph{Context--target conditional prediction.}
For each prediction episode, we sample a context set
\begin{equation}
\mathcal{C}=\{(\mathbf{x}_i^{\mathcal{C}},\mathbf{y}_i^{\mathcal{C}})\}_{i=1}^{n_c}\subseteq \mathcal{D}_{\mathcal{H}},
\end{equation}
and a target set
\begin{equation}
\mathcal{T}=\{(\mathbf{x}_j^{\mathcal{T}},\mathbf{y}_j^{\mathcal{T}})\}_{j=1}^{n_t},
\end{equation}
where $n_c$ and $n_t$ denote the number of context and target points.
Our objective is to model the conditional predictive distribution
\begin{equation}
p_\Theta\!\left(\mathbf{y}^{\mathcal{T}} \mid \mathbf{x}^{\mathcal{T}}, \mathcal{C}\right),
\end{equation}
with emphasis on target inputs in extreme regimes (typically $\mathcal{D}_{\mathcal{T}}^{e}$).
We use the term \emph{phase transition} to refer to abrupt changes in the conditional mapping $p(\mathbf{y}\mid\mathbf{x})$ across regimes, which makes context selection and weighting crucial under extreme scenarios.

\subsection{Conditional Neural Processes Preliminaries}
Conditional Neural Processes (CNPs) \citep{garnelo2018conditional} perform amortized conditional density estimation by encoding a context set and decoding a predictive distribution at target inputs.
Given a context set $\mathcal{C}=\{(\mathbf{x}_i^{\mathcal{C}},\mathbf{y}_i^{\mathcal{C}})\}_{i=1}^{n_c}$, an encoder network $h_\theta$ maps each context pair to a representation vector
\begin{equation}
\mathbf{r}_i = h_{\theta}\!\left(\mathbf{x}_i^{\mathcal{C}}, \mathbf{y}_i^{\mathcal{C}}\right), \qquad
h_{\theta}:\mathcal{X}\times\mathcal{Y}\rightarrow\mathbb{R}^{d_r},
\end{equation}
where $d_r$ is the representation dimension.
Standard CNP aggregates context representations using a permutation-invariant mean operator
\begin{equation}
\mathbf{r}=\frac{1}{n_c}\sum_{i=1}^{n_c}\mathbf{r}_i.
\end{equation}
A decoder network $g_\theta$ then predicts distribution parameters for each target input $\mathbf{x}_j^{\mathcal{T}}$:
\begin{equation}
\boldsymbol{\phi}_j = g_{\theta}\!\left(\mathbf{x}_j^{\mathcal{T}}, \mathbf{r}\right), \qquad
g_\theta:\mathcal{X}\times\mathbb{R}^{d_r}\rightarrow\Phi.
\end{equation}
For regression, we instantiate $p_\Theta(\mathbf{y}\mid\boldsymbol{\phi})$ as a Gaussian distribution and let $\boldsymbol{\phi}_j=(\boldsymbol{\mu}_j,\boldsymbol{\sigma}_j^2)$ parameterize its mean and (diagonal) variance.
The model is trained by minimizing the negative log-likelihood on targets sampled from the same training pool:
\begin{equation}
\mathcal{L}_{\textsc{cnp}}(\theta)
=
-\mathbb{E}_{\mathcal{C},\mathcal{T}\subseteq \mathcal{D}_{\mathcal{H}}}
\left[\sum_{j=1}^{n_t}\log p_{\theta}\!\left(\mathbf{y}_j^{\mathcal{T}} \mid \boldsymbol{\phi}_j\right)\right].
\end{equation}

While effective under typical conditions, the uniform mean aggregation can dilute the influence of those context points that are most relevant to an extreme target query, especially when $p(\mathbf{y}\mid\mathbf{x})$ shifts across regimes.
This motivates target-conditioned context weighting.

\subsection{Adaptive Weighting Mechanism}
Standard CNPs use uniform mean aggregation, implicitly assuming that all context points contribute equally to predicting any target.
This assumption is problematic under extreme events, where the conditional relationship $p(\mathbf{y}\mid \mathbf{x})$ may shift and only a small subset of historical contexts is informative for a given target query.
AdaCNP addresses this by introducing a target-conditioned adaptive layer that assigns relevance weights to context points.

\begin{figure}[t]
\centering
\small
\begin{minipage}[t]{0.48\textwidth}
\begin{algorithm}[H]
\small
\caption{AdaCNP Training (End-to-End)}
\label{alg:train}
\begin{algorithmic}[1]
\REQUIRE Historical dataset $\mathcal{D}_{\mathcal{H}}$;
encoder $h_{\theta}$, decoder $g_{\theta}$; embedding network $\phi_{\omega}$; adaptive layer $f_{\psi}$;
temperature $\tau>0$
\STATE Initialize parameters $\theta,\omega,\psi$
\FOR{$iter = 1$ to $N$}
    \STATE Sample context set $\mathcal{C}=\{(\mathbf{x}_i^{\mathcal{C}},\mathbf{y}_i^{\mathcal{C}})\}_{i=1}^{n_c}\subseteq \mathcal{D}_{\mathcal{H}}$
    \STATE Sample target set $\mathcal{T}=\{(\mathbf{x}_j^{\mathcal{T}},\mathbf{y}_j^{\mathcal{T}})\}_{j=1}^{n_t}\subseteq \mathcal{D}_{\mathcal{H}}$
    \STATE Encode contexts: $\mathbf{r}_i \leftarrow h_{\theta}(\mathbf{x}_i^{\mathcal{C}},\mathbf{y}_i^{\mathcal{C}})$ for $i=1,\dots,n_c$
    \STATE Embed inputs: $\mathbf{e}_i^{\mathcal{C}} \leftarrow \phi_{\omega}(\mathbf{x}_i^{\mathcal{C}})$ for $i=1,\dots,n_c$
    \STATE Embed inputs: $\mathbf{e}_j^{\mathcal{T}} \leftarrow \phi_{\omega}(\mathbf{x}_j^{\mathcal{T}})$ for $j=1,\dots,n_t$
    \FOR{$j = 1$ to $n_t$}
        \STATE Scores: $s_{ij}\leftarrow f_{\psi}(\mathbf{e}_i^{\mathcal{C}},\mathbf{e}_j^{\mathcal{T}})$ for all $i$
        \STATE Weights: $w_{ij}\leftarrow \frac{\exp(s_{ij}/\tau)}{\sum_{i'=1}^{n_c}\exp(s_{i'j}/\tau)}$ for all $i$
        \STATE Aggregate: $\mathbf{r}_j \leftarrow \sum_{i=1}^{n_c} w_{ij}\,\mathbf{r}_i$
        \STATE Decode: $\boldsymbol{\phi}_j \leftarrow g_{\theta}(\mathbf{x}_j^{\mathcal{T}}, \mathbf{r}_j)$
    \ENDFOR
    \STATE Loss: $\mathcal{L}_{\textsc{adacnp}} \leftarrow -\sum_{j=1}^{n_t}\log p_{\theta}(\mathbf{y}_j^{\mathcal{T}}\mid \boldsymbol{\phi}_j)$
    \STATE Update: $(\theta,\omega,\psi)\leftarrow (\theta,\omega,\psi)-\eta\nabla_{(\theta,\omega,\psi)}\mathcal{L}_{\textsc{adacnp}}$
    
\ENDFOR
\STATE \textbf{Return} trained parameters $\theta,\omega,\psi$
\end{algorithmic}
\end{algorithm}
\end{minipage}%
\hfill
\begin{minipage}[t]{0.48\textwidth}
\begin{algorithm}[H]
\small
\caption{Inference with Trained AdaCNP}
\label{alg:inference}
\begin{algorithmic}[1]
\REQUIRE Trained $h_{\theta}$, $g_{\theta}$, $\phi_{\omega}$, $f_{\psi}$;
historical dataset $\mathcal{D}_{\mathcal{H}}$; target dataset $\mathcal{D}_{\mathcal{T}}$;
temperature $\tau>0$
\FOR{each target input $\mathbf{x}_j^{\mathcal{T}} \in \mathcal{D}_{\mathcal{T}}$}
    \STATE Sample context set $\mathcal{C}=\{(\mathbf{x}_i^{\mathcal{C}},\mathbf{y}_i^{\mathcal{C}})\}_{i=1}^{n_c}\subseteq \mathcal{D}_{\mathcal{H}}$
    \STATE Encode contexts: $\mathbf{r}_i \leftarrow h_{\theta}(\mathbf{x}_i^{\mathcal{C}},\mathbf{y}_i^{\mathcal{C}})$ for $i=1,\dots,n_c$
    \STATE Embed contexts: $\mathbf{e}_i^{\mathcal{C}} \leftarrow \phi_{\omega}(\mathbf{x}_i^{\mathcal{C}})$ for $i=1,\dots,n_c$
    \STATE Embed target: $\mathbf{e}_j^{\mathcal{T}} \leftarrow \phi_{\omega}(\mathbf{x}_j^{\mathcal{T}})$
    \STATE Scores: $s_{ij}\leftarrow f_{\psi}(\mathbf{e}_i^{\mathcal{C}},\mathbf{e}_j^{\mathcal{T}})$ for all $i$
    \STATE Weights: $w_{ij}\leftarrow \frac{\exp(s_{ij}/\tau)}{\sum_{i'=1}^{n_c}\exp(s_{i'j}/\tau)}$ for all $i$
    \STATE Aggregate: $\mathbf{r}_j \leftarrow \sum_{i=1}^{n_c} w_{ij}\,\mathbf{r}_i$
    \STATE Decode: $\boldsymbol{\phi}_j \leftarrow g_{\theta}(\mathbf{x}_j^{\mathcal{T}}, \mathbf{r}_j)$
    \STATE Output predictive distribution $p_{\theta}(\mathbf{y}\mid \boldsymbol{\phi}_j)$ (e.g., $\mathcal{N}(\boldsymbol{\mu}_j,\boldsymbol{\sigma}_j^2)$)
\ENDFOR
\end{algorithmic}
\end{algorithm}
\end{minipage}
\end{figure}

\paragraph{Embedding network.}
Given a context set $\mathcal{C}=\{(\mathbf{x}_i^{\mathcal{C}},\mathbf{y}_i^{\mathcal{C}})\}_{i=1}^{n_c}$ and target inputs $\{\mathbf{x}_j^{\mathcal{T}}\}_{j=1}^{n_t}$, we first embed inputs into a learned embedding space via
\begin{equation}
\phi_\omega:\mathcal{X}\rightarrow \mathbb{R}^{d_e},
\end{equation}
where $d_e$ denotes the embedding dimension.
The context and target embeddings are computed as
\begin{equation}
\mathbf{e}_i^{\mathcal{C}}=\phi_\omega(\mathbf{x}_i^{\mathcal{C}}),\qquad
\mathbf{e}_j^{\mathcal{T}}=\phi_\omega(\mathbf{x}_j^{\mathcal{T}}).
\end{equation}

\paragraph{Adaptive scoring layer $f_\psi$.}
We introduce an adaptive scoring function
\begin{equation}
f_\psi:\mathbb{R}^{d_e}\times \mathbb{R}^{d_e}\rightarrow \mathbb{R},
\end{equation}
which takes a context embedding and a target embedding as input and outputs a scalar relevance score:
\begin{equation}
s_{ij}=f_\psi(\mathbf{e}_i^{\mathcal{C}},\mathbf{e}_j^{\mathcal{T}}).
\end{equation}
The function $f_\psi$ can be instantiated as a lightweight MLP or a parametric similarity module; its role is to learn a task-specific notion of relevance between contexts and targets.

\paragraph{Normalized target-conditioned weights.}
We convert scores into nonnegative, normalized weights via a softmax over context indices:
\begin{equation}
w_{ij}
=
\frac{\exp(s_{ij}/\tau)}{\sum_{i'=1}^{n_c}\exp(s_{i'j}/\tau)},
\qquad
\sum_{i=1}^{n_c} w_{ij}=1,\;\; w_{ij}\ge 0,
\end{equation}
where $\tau>0$ is a temperature parameter controlling the concentration of weights.

\paragraph{Weighted aggregation and decoding.}
Let $\mathbf{r}_i=h_\theta(\mathbf{x}_i^{\mathcal{C}},\mathbf{y}_i^{\mathcal{C}})\in\mathbb{R}^{d_r}$ be the CNP context representations produced by the encoder.
For each target query $j$, AdaCNP forms a target-specific aggregated representation by a weighted sum:
\begin{equation}
\mathbf{r}_j=\sum_{i=1}^{n_c} w_{ij}\,\mathbf{r}_i
=
\sum_{i=1}^{n_c} w_{ij}\,h_{\theta}\!\left(\mathbf{x}_i^{\mathcal{C}}, \mathbf{y}_i^{\mathcal{C}}\right).
\end{equation}
The decoder predicts the target distribution parameters using the same interface as CNP:
\begin{equation}
\boldsymbol{\phi}_j=g_\theta(\mathbf{x}_j^{\mathcal{T}},\mathbf{r}_j).
\end{equation}
This construction is permutation-invariant with respect to the ordering of context points and enables query-specific conditioning that is critical under extreme scenarios.

\subsection{Training Procedure}
To train the proposed AdaCNP, a set of context points and target points is sampled from the historical dataset in each epoch. At each iteration, we sample a context set $\mathcal{C}\subseteq \mathcal{D}_{\mathcal{H}}$ and a target set $\mathcal{T}\subseteq \mathcal{D}_{\mathcal{H}}$. The context and target features are passed through the adaptive layer, which computes relevance weights based on the similarity between their embeddings. These weights are used to aggregate the context representations, which are then passed through the decoder to predict the distribution parameters for the target. The NLL loss is computed between the predicted and true target values, and the model parameters are updated using backpropagation.
During inference, contexts are sampled exclusively from $\mathcal{D}_{\mathcal{H}}$ to prevent information leakage from the target split.


\subsection{Implications and Extensions}
The AdaCNP framework provides a robust and efficient solution for rapidly adapting to new and challenging environments without requiring complete retraining. While it is tailored for load forecasting, the adaptability of AdaCNP makes it well-suited for a wide range of domains where distribution shifts are common and data is often scarce. This includes applications such as climate anomaly prediction, financial risk assessment, where the ability to accurately predict extreme events is crucial.

One of the key strengths of AdaCNP is its ability to preserve important qualities such as interpretability, computational efficiency, and uncertainty quantification. These features not only enhance the model’s reliability but also make it an invaluable tool for researchers and practitioners who need to make data-driven decisions in high-stakes, data-limited forecasting scenarios. By focusing on relevant context points and adjusting to distributional shifts dynamically, AdaCNP provides a flexible, efficient, and interpretable solution for real-world predictive challenges.

\section{Experimental Results}

\subsection{Experiment 1: 1D Regression with Phase Transition}

To systematically investigate forecasting challenges, we construct a toy model that captures the essential features of extreme-event forecasting. Our model considers a system where an output variable $y$ depends on input parameters $x$, exhibiting a phase transition beyond a critical threshold. This transition creates distinct behavioral regimes that mirror real-world phenomena - a "normal" regime where standard statistical assumptions hold, and an "extreme" regime governed by fundamentally different dynamics.
For a fixed threshold value $x_c \in \mathbb{R}$, we define the conditional probability distribution of $y$ given $x$ as:
$$
y|x \sim \begin{cases}
\mathcal{N}(\mu_1(x), \sigma_1^2(x)) & \text{for } x \leq x_c \\
\mathcal{N}(\mu_2(x), \sigma_2^2(x)) & \text{for } x > x_c
\end{cases}
$$
where $x, y \in \mathbb{R}$ represent the input parameter and output variable, respectively. The functions $\mu_1(x), \mu_2(x), \sigma_1^2(x), \sigma_2^2(x)$ may exhibit discontinuity at $x = x_c$, modeling abrupt transitions observed in real-world systems.

In our implementation, we set $x_c = 1$ and  choose $\mu_1(x)$ as a linear function with randomly sampled slope and intercept and $\mu_2(x)$ as a sinusoidal function with random frequency and phase that preserves continuity at the transition point. The variance $\sigma_1^2$ and $\sigma_2^2$ are randomly generated in each regime, mimicking the noise level transition for real-world data. This setup provides a controlled environment to study prediction challenges under regime shifts while remaining mathematically tractable.
The model is trained using a Negative Log-Likelihood (NLL) loss function.

We evaluate the performance of AdaCNP against the following baselines: (1) Conditional Neural Processes (CNP); (2) Attentive Neural Processes (ANP); (3) Gaussian Process (GP); (4) Neural Processes (NP).  
Their performances are evaluated using different numbers of context points to demonstrate how the model's performance improves with the amount of available context data.

\begin{table}[b]
\caption{NLL and MSE on 1-D regression with phase transition.}
\label{sample-table-toy}
\begin{center}
\begin{small}
\begin{sc}
\begin{tabular}{@{}l@{\hspace{8pt}}c@{\hspace{8pt}}c@{\hspace{8pt}}c@{\hspace{8pt}}c@{\hspace{8pt}}c@{}}
\toprule
  & AdaCNP    & CNP       & ANP       & GP       & NP \\
\midrule
MSE(\%)   & \textbf{0.895}  & 0.904  & 1.104  & 98.691 & 1.099  \\ 
NLL   & \textbf{-0.944} & -0.919 & -0.881 & 1.150 & -0.846 \\
\bottomrule
\end{tabular}
\end{sc}
\end{small}
\end{center} 
\end{table}


\begin{figure}[t]
    \centering
    \includegraphics[scale=0.4]{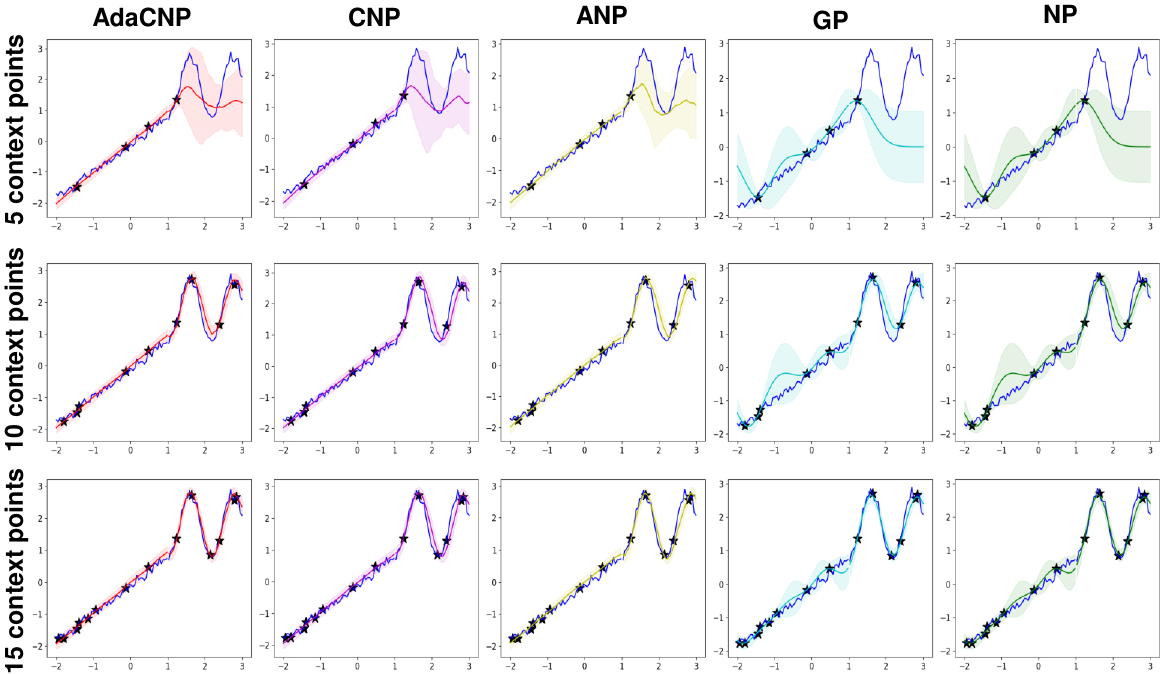}
    \caption{Regression results on a 1-D curve with phase transition (blue line) using 5, 10, and 15 context points (black stars). The color curves, as well as the surrounding shallow range, show the predicted mean and variance using AdaCNP, CNP, ANP, GP, and NP.}  
    \vspace{-1.5em}
    \label{fig2}  
\end{figure}

Table \ref{sample-table-toy}, AdaCNP achieves the best performance on both evaluation metrics. AdaCNP outperforms CNP by  1.0\%, ANP by 18.9\%, and NP by 18.5\%, and they show a dramatic improvement over GP. Similarly, AdaCNP improves NLL upon CNP by 2.7\%, ANP by 7.1\%, and NP by 11.6\%. These results highlight AdaCNP’s advantage in both prediction accuracy and uncertainty estimation compared to baselines.

Figure~\ref{fig2} also demonstrates the superiority of our model.
AdaCNP outperforms the other models, particularly in the extreme regime where the phase transition occurs. As the number of context points increases, AdaCNP consistently delivers more accurate predictions with tighter confidence intervals around the predicted mean. This demonstrates its ability to adapt to distribution shifts during extreme events effectively.
While CNP and ANP perform reasonably well, they lag behind AdaCNP in accuracy. The absence of an adaptive weighting mechanism in these models likely limits their performance in the extreme regime.
Both GP and NP perform adequately in normal conditions but exhibit wider confidence intervals, indicating higher uncertainty, especially when faced with extreme values.
These results underscore the effectiveness of AdaCNP in managing distribution shifts. 
While all models perform relatively well given sufficient context points, performance deviations become more significant in extreme load forecasting with limited real-world data. This highlights the strength of AdaCNP's adaptive layer, allowing it to quickly adjust to new, unseen extreme scenarios—making it a valuable tool for forecasting in high-risk, data-scarce environments.

\subsection{Experiment 2: Extreme Load Forecasting }

\begin{figure}[t]
\centering

\begin{minipage}[c]{0.48\textwidth}
    \begin{minipage}[c]{\textwidth}
        \includegraphics[width=\textwidth]{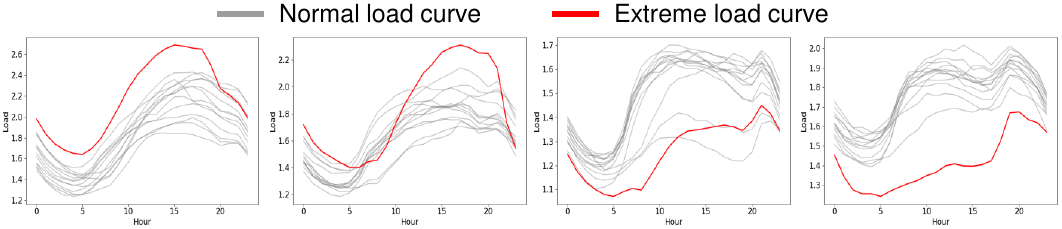}
        \caption{Illustrative example of 4 extreme load curves identified in the PJM dataset. The grey curves represent normal load patterns, while the red curves are the identified extreme samples.}
        \label{fig3}
    \end{minipage}


    \begin{minipage}[c]{\textwidth}
        \includegraphics[width=\textwidth]{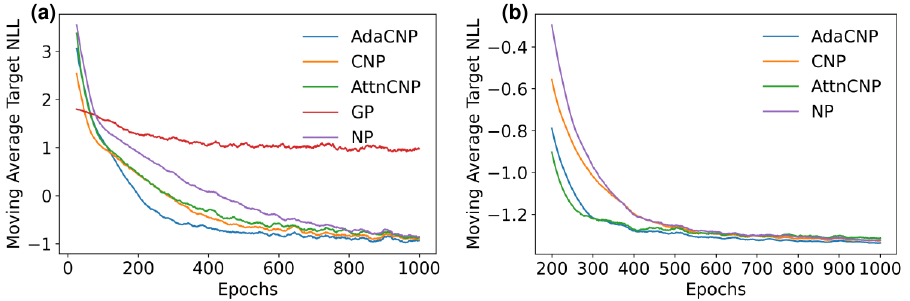}
        \caption{Moving average of the training NLL loss for the load forecasting task on: (a) PJM dataset (left) and (b) ISO-NE dataset (right).}
        \label{loss_curve}
    \end{minipage}
\end{minipage}
\end{figure}

We present an experiment to evaluate the effectiveness of the AdaCNP model for forecasting extreme electricity consumption. Our primary objective is to demonstrate the model's ability to accurately predict extreme load conditions where the data is sparse and highly variable. To this end, we partition the dataset into two subsets: the \textit{Normal Set} (non-extreme days) and the \textit{Extreme Set} (extreme days), with extreme samples identified using a 3-$\sigma$ threshold to filter out outliers. Specifically, we employ the Dynamic Time Warping (DTW) approach with a sliding window. We compare the load curve for each day with those from the 7 days preceding and following it. Days exhibiting significantly higher differences from the rest of the window (measured by both magnitude and shape) are classified as extreme events. The criterion for an extreme event is that the mean DTW distance exceeds three standard deviations from the average distance in the window. An illustrative example of extreme load curves is shown in Fig. \ref{fig3}.


Once extreme days are identified, the dataset is split into the \textit{Normal Set} and \textit{Extreme Set}, and partitioned into training and test sets. The test set for extreme days is kept separate to evaluate how well the models generalize to rare events. We evaluate AdaCNP using two real-world datasets. 

(a) PJM dataset: An 8-year electrical load data from PJM Interconnection LLC, which is a regional transmission organization in the United States. The task is to predict the electricity consumption for the next 24 hours. For each day, we input the past day’s electrical load and temperature, the next day’s temperature forecast, and additional features such as non-linear functions of the temperatures, binary indicators of weekends or holidays, and yearly features. 

(b) ISO-NE dataset: A North American utility's load and temperature data at the one-hour resolution, covering the time range between January 1st, 1985 and October 12th, 1992. Input features include past load and weather information, and one-hot encodings for season, weekday/weekend, and holiday/non-holiday, which help to capture temporal patterns.
  
To model uncertainty in the forecast, AdaCNP outputs both the mean $\mu$ and variance $\sigma$ for each of the 24 hourly time steps.  We evaluate the performance of AdaCNP against the same baseline models as experiment 1. 
Each model is trained and evaluated on the same dataset, with a focus on assessing performance during extreme load conditions.

\begin{table*}[t] 
\caption{NLL and MSE on PJM dataset and ISO-NE dataset.}
\vspace{-1em}
\label{sample-table}
\begin{center}
\begin{footnotesize}
\begin{sc}
\begin{tabular}{@{}l@{\hspace{8pt}}c@{\hspace{8pt}}c@{\hspace{8pt}}c@{\hspace{8pt}}c@{\hspace{8pt}}c@{}}
\toprule
PJM & AdaCNP    & CNP       & ANP       & GP       & NP \\
\midrule
MSE (\%)   & \textbf{0.95 $\pm$ 0.09} & 0.99 $\pm$ 0.11 & 1.44  $\pm$ 0.16  & 75.72 $\pm$ 6.40 & 1.21 $\pm$ 0.11 \\ 
NLL        & \textbf{-0.92 $\pm$ 3.4\%} & -0.89 $\pm$ 3.8\% & -0.80 $\pm$ 3.2\% & 0.97 $\pm$ 4.1\% & -0.80 $\pm$ 3.6\% \\ 
Pinball    & \textbf{0.02 $\pm$ 0.09\%} & 0.02 $\pm$ 0.10\% & 0.03 $\pm$ 0.11\% & 0.19  $\pm$ 0.82\% & 0.03 $\pm$ 0.10\% \\   

\bottomrule
\end{tabular}
\end{sc}
\end{footnotesize}
\end{center}
\begin{center}
\begin{footnotesize}
\begin{sc}
\begin{tabular}{@{}l@{\hspace{8pt}}c@{\hspace{8pt}}c@{\hspace{8pt}}c@{\hspace{8pt}}c@{\hspace{8pt}}c@{}}
\toprule
ISO-NE & AdaCNP    & CNP       & ANP       & GP       & NP \\
\midrule
MSE (\%)   & \textbf{0.10 $\pm$ 0.01} & 0.11 $\pm$ 0.01  & 0.11  $\pm$ 0.01  & 3.39 $\pm$ 0.01 & 0.113 $\pm$ 0.28 \\ 
NLL        & \textbf{-1.33 $\pm$ 4.6\%} & -1.32 $\pm$ 6.6\% & -1.32 $\pm$ 6.1\% & 0.818 $\pm$ 8.1\% &  -1.32 $\pm$ 6.7\% \\ 
Pinball    & \textbf{0.012 $\pm$ 0.02\%} & 0.013 $\pm$ 0.03\% & 0.013 $\pm$ 0.02\% & 0.013 $\pm$ 0.03\% & 0.099 $\pm$ 0.02\% \\   
\bottomrule
\end{tabular}
\end{sc}
\end{footnotesize}
\end{center} 
\end{table*}  

\begin{figure*}[t]
    \centering
    \includegraphics[width=\textwidth]{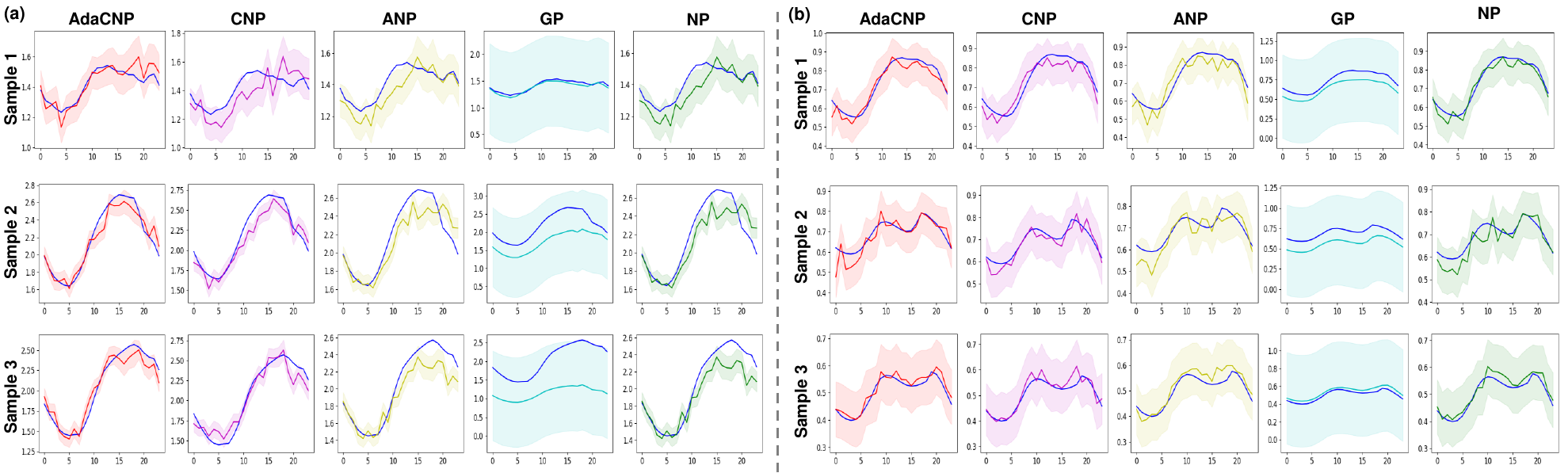}
    \caption{Predictions of the AdaCNP, CNP, AttnCNP, GP and NP models on three extreme load curves from the test set of unseen extreme samples over a 24-hour forecasting horizon from (a) PJM dataset (left) and (b) ISO-NE dataset (right). The ground truth (blue) is the actual electricity consumption and the colored curves represent the mean predictions of each model. The shaded regions around each curve are confidence intervals, denoted by the model's prediction variance ($\mu \pm \sigma$).}
    \label{load_forecast}
\end{figure*}

As shown in Fig. \ref{loss_curve}, AdaCNP consistently outperforms the baseline models across all epochs. For clarity of visualization in the ISO-NE dataset subfigure, we exclude the GP model results due to their substantially higher loss values. From Table~\ref{sample-table}, on PJM dataset, AdaCNP achieves state-of-the-art performance, reducing MSE by 17\% compared to ANP and NP, highlighting its robustness to extreme variability where traditional methods fail. In uncertainty quantification, AdaCNP’s negative log-likelihood also surpasses all baseline models, demonstrating sharper calibration despite sparse training data.
In terms of Pinball loss, which evaluates the accuracy of quantile predictions, AdaCNP consistently outperforms all baselines across both datasets. On PJM dataset, it achieves a 1.3\% reduction compared to the next best model CNP.
On the ISO-NE dataset, AdaCNP again attains the best MSE and NLL, improving upon the next strongest baseline CNP and NP by over 22\% in MSE and providing sharper calibration than any competing model. For Pinbal loss, it improves by 2.4\%, indicating more precise and well-calibrated probabilistic forecasts. Overall, these results highlight AdaCNP’s ability to adapt effectively to differing load patterns and feature sets, while maintaining superior performance in uncertainty estimation for both typical and extreme load conditions.

Further evaluation was conducted on three example extreme load consumption curves identified from two datasets. As shown in Fig.\ref{load_forecast}, AdaCNP provides the most accurate predictions, closely tracking the ground truth across the 24-hour forecast period. The confidence interval is also appropriately tight, reflecting the model’s high certainty in predicting extreme consumption patterns.
In contrast, CNP and ANP perform relatively well but show noticeable deviations from the ground truth, especially during periods of significant fluctuations. Their confidence intervals are wider, indicating higher uncertainty in extreme load predictions. NP shows a similar trend, with its predictions deviating from the true consumption pattern, especially during the more extreme load variations. GP struggles the most with extreme consumption patterns, exhibiting substantial deviations from the ground truth. This is reflected both in the mean prediction curve and the wider confidence intervals, indicating the model's inability to model non-stationary extreme load conditions. 



\section{Conclusion}
In this work, we introduced AdaCNP, a novel framework designed for extreme load forecasting that effectively handles distribution shifts and sparse data. Through both a carefully designed toy model with phase transitions and real-world extreme load forecasting experiments, we demonstrated AdaCNP's capability to capture abrupt behavioral changes and adapt to different regimes. The model achieves the lowest MSE and NLL loss in real-world extreme load forecasting tasks, highlighting its effectiveness in high-risk, data-scarce environments. Looking forward, AdaCNP's success in load forecasting suggests promising applications in broader power system operations, particularly for critical infrastructure where reliable predictions under extreme conditions are essential.

Our work contributes to the field by demonstrating that embedding-based similarity learning provides a powerful alternative to traditional out-of-distribution (OOD) approaches, particularly in extreme-event scenarios where labeled examples are scarce. Unlike conventional OOD methods that require learning invariant representations across predefined domains with substantial data, AdaCNP dynamically adapts through a learned similarity space without needing explicit domain boundaries. By constructing a unified embedding where relevance across different regimes is learned directly, AdaCNP can effectively reweight historical patterns based on their utility for the current prediction target. This enables robust forecasting under severe distribution shifts while preserving model simplicity and interpretability. Future work could explore extending this framework to other domains facing similar challenges of uncertainty and limited data, such as climate anomaly detection, financial risk analysis, or rare disease prognosis.

\bibliographystyle{named}
\bibliography{ijcai26}

@inproceedings{bhatia2021exgan,
  title={Exgan: Adversarial generation of extreme samples},
  author={Bhatia, Siddharth and Jain, Arjit and Hooi, Bryan},
  booktitle={Proceedings of the AAAI Conference on Artificial Intelligence},
  volume={35}, 
  pages={6750--6758},
  year={2021}
}

@article{chen2004load,
  title={Load forecasting using support vector machines: A study on EUNITE competition 2001},
  author={Chen, Bo-Juen and Chang, Ming-Wei and others},
  journal={IEEE transactions on power systems},
  volume={19},
  number={4},
  pages={1821--1830},
  year={2004},
  publisher={IEEE}
}

@article{hippert2001neural,
  title={Neural networks for short-term load forecasting: A review and evaluation},
  author={Hippert, Henrique Steinherz and Pedreira, Carlos Eduardo and Souza, Reinaldo Castro},
  journal={IEEE Transactions on power systems},
  volume={16},
  number={1},
  pages={44--55},
  year={2001},
  publisher={IEEE}
}

@article{kong2017short,
  title={Short-term residential load forecasting based on LSTM recurrent neural network},
  author={Kong, Weicong and Dong, Zhao Yang and Jia, Youwei and Hill, David J and Xu, Yan and Zhang, Yuan},
  journal={IEEE transactions on smart grid},
  volume={10},
  number={1},
  pages={841--851},
  year={2017},
  publisher={IEEE}
}

@article{wang2022transformer,
  title={A transformer-based method of multienergy load forecasting in integrated energy system},
  author={Wang, Chen and Wang, Ying and Ding, Zhetong and Zheng, Tao and Hu, Jiangyi and Zhang, Kaifeng},
  journal={IEEE Transactions on Smart Grid},
  volume={13},
  number={4},
  pages={2703--2714},
  year={2022},
  publisher={IEEE}
}

@inproceedings{liu2023sadi,
  title={Sadi: A self-adaptive decomposed interpretable framework for electric load forecasting under extreme events},
  author={Liu, Hengbo and Ma, Ziqing and Yang, Linxiao and Zhou, Tian and Xia, Rui and Wang, Yi and Wen, Qingsong and Sun, Liang},
  booktitle={ICASSP 2023-2023 IEEE International Conference on Acoustics, Speech and Signal Processing (ICASSP)},
  pages={1--5},
  year={2023},
  organization={IEEE}
}

@inproceedings{garnelo2018conditional,
  title={Conditional neural processes},
  author={Garnelo, Marta and Rosenbaum, Dan and Maddison, Christopher and Ramalho, Tiago and Saxton, David and Shanahan, Murray and Teh, Yee Whye and Rezende, Danilo and Eslami, SM Ali},
  booktitle={International conference on machine learning},
  pages={1704--1713},
  year={2018},
  organization={PMLR}
}

@article{liu2015probabilistic,
  title={Probabilistic load forecasting via quantile regression averaging on sister forecasts},
  author={Liu, Bidong and Nowotarski, Jakub and Hong, Tao and Weron, Rafa{\l}},
  journal={IEEE Transactions on Smart Grid},
  volume={8},
  number={2},
  pages={730--737},
  year={2015},
  publisher={IEEE}
}

@article{haben2016hybrid,
  title={A hybrid model of kernel density estimation and quantile regression for GEFCom2014 probabilistic load forecasting},
  author={Haben, Stephen and Giasemidis, Georgios},
  journal={International Journal of Forecasting},
  volume={32},
  number={3},
  pages={1017--1022},
  year={2016},
  publisher={Elsevier}
}

@article{wan2021adaptive,
  title={An adaptive ensemble data driven approach for nonparametric probabilistic forecasting of electricity load},
  author={Wan, Can and Cao, Zhaojing and Lee, Wei-Jen and Song, Yonghua and Ju, Ping},
  journal={IEEE Transactions on Smart Grid},
  volume={12},
  number={6},
  pages={5396--5408},
  year={2021},
  publisher={IEEE}
}

@article{wen2022continuous,
  title={Continuous and distribution-free probabilistic wind power forecasting: A conditional normalizing flow approach},
  author={Wen, Honglin and Pinson, Pierre and Ma, Jinghuan and Gu, Jie and Jin, Zhijian},
  journal={IEEE Transactions on Sustainable Energy},
  volume={13},
  number={4},
  pages={2250--2263},
  year={2022},
  publisher={IEEE}
}

@book{haben2023core,
  title={Core Concepts and Methods in Load Forecasting: With Applications in Distribution Networks},
  author={Haben, Stephen and Voss, Marcus and Holderbaum, William},
  year={2023},
  publisher={Springer Nature}
}

@article{yang2019bayesian,
  title={Bayesian deep learning-based probabilistic load forecasting in smart grids},
  author={Yang, Yandong and Li, Wei and Gulliver, T Aaron and Li, Shufang},
  journal={IEEE Transactions on Industrial Informatics},
  volume={16},
  number={7},
  pages={4703--4713},
  year={2019},
  publisher={IEEE}
}

@article{hu2022black,
  title={Black swan event small-sample transfer learning (BEST-L) and its case study on electrical power prediction in COVID-19},
  author={Hu, Chenxi and Zhang, Jun and Yuan, Hongxia and Gao, Tianlu and Jiang, Huaiguang and Yan, Jing and Gao, David Wenzhong and Wang, Fei-Yue},
  journal={Applied Energy},
  volume={309},
  pages={118458},
  year={2022},
  publisher={Elsevier}
}

@inproceedings{laptev2017time,
  title={Time-series extreme event forecasting with neural networks at uber},
  author={Laptev, Nikolay and Yosinski, Jason and Li, Li Erran and Smyl, Slawek},
  booktitle={International conference on machine learning},
  volume={34},
  pages={1--5},
  year={2017},
  organization={sn}
}

@article{gu2021imbalance,
  title={An imbalance modified convolutional neural network with incremental learning for chemical fault diagnosis},
  author={Gu, Xiaohua and Zhao, Yanli and Yang, Guang and Li, Lusi},
  journal={IEEE Transactions on Industrial Informatics},
  volume={18},
  number={6},
  pages={3630--3639},
  year={2021},
  publisher={IEEE}
}

@article{kim2019attentive,
  title={Attentive neural processes},
  author={Kim, Hyunjik and Mnih, Andriy and Schwarz, Jonathan and Garnelo, Marta and Eslami, Ali and Rosenbaum, Dan and Vinyals, Oriol and Teh, Yee Whye},
  journal={arXiv preprint arXiv:1901.05761},
  year={2019}
}

@article{garnelo2018neural,
  title={Neural processes},
  author={Garnelo, Marta and Schwarz, Jonathan and Rosenbaum, Dan and Viola, Fabio and Rezende, Danilo J and Eslami, SM and Teh, Yee Whye},
  journal={arXiv preprint arXiv:1807.01622},
  year={2018}
}

@article{li2019use,
  title={The use of extreme value theory for forecasting long-term substation maximum electricity demand},
  author={Li, Yun and Jones, Ben},
  journal={IEEE Transactions on Power Systems},
  volume={35},
  number={1},
  pages={128--139},
  year={2019},
  publisher={IEEE}
}

@book{haan2006extreme,
  title={Extreme value theory: an introduction},
  author={Haan, Laurens and Ferreira, Ana},
  volume={3},
  year={2006},
  publisher={Springer}
}

@article{liu2021towards,
  title={Towards out-of-distribution generalization: A survey},
  author={Liu, Jiashuo and Shen, Zheyan and He, Yue and Zhang, Xingxuan and Xu, Renzhe and Yu, Han and Cui, Peng},
  journal={arXiv preprint arXiv:2108.13624},
  year={2021}
}

@article{arjovsky2019invariant,
  title={Invariant risk minimization},
  author={Arjovsky, Martin and Bottou, L{\'e}on and Gulrajani, Ishaan and Lopez-Paz, David},
  journal={arXiv preprint arXiv:1907.02893},
  year={2019}
}

@article{gama2014survey,
  title={A survey on concept drift adaptation},
  author={Gama, Jo{\~a}o and {\v{Z}}liobait{\.e}, Indr{\.e} and Bifet, Albert and Pechenizkiy, Mykola and Bouchachia, Abdelhamid},
  journal={ACM computing surveys (CSUR)},
  volume={46},
  number={4},
  pages={1--37},
  year={2014},
  publisher={ACM New York, NY, USA}
}

@article{WANG2018135,
title = {Deep visual domain adaptation: A survey},
journal = {Neurocomputing},
volume = {312},
pages = {135-153},
year = {2018},
issn = {0925-2312},
doi = {https://doi.org/10.1016/j.neucom.2018.05.083},
url = {https://www.sciencedirect.com/science/article/pii/S0925231218306684},
author = {Mei Wang and Weihong Deng}
}

@ARTICLE{zhuang2020comprehensive,
  author={Zhuang, Fuzhen and Qi, Zhiyuan and Duan, Keyu and Xi, Dongbo and Zhu, Yongchun and Zhu, Hengshu and Xiong, Hui and He, Qing},
  journal={Proceedings of the IEEE}, 
  title={A Comprehensive Survey on Transfer Learning}, 
  year={2021},
  volume={109},
  number={1},
  pages={43-76},
  doi={10.1109/JPROC.2020.3004555}}

@article{fekri2021deep,
  title={Deep learning for load forecasting with smart meter data: Online Adaptive Recurrent Neural Network},
  author={Fekri, Mohammad Navid and Patel, Harsh and Grolinger, Katarina and Sharma, Vinay},
  journal={Applied Energy},
  volume={282},
  pages={116177},
  year={2021},
  publisher={Elsevier}
}

@article{bayram2023lstm,
  title={DA-LSTM: A dynamic drift-adaptive learning framework for interval load forecasting with LSTM networks},
  author={Bayram, Firas and Aupke, Phil and Ahmed, Bestoun S and Kassler, Andreas and Theocharis, Andreas and Forsman, Jonas},
  journal={Engineering Applications of Artificial Intelligence},
  volume={123},
  pages={106480},
  year={2023},
  publisher={Elsevier}
}

@misc{
zhu2024learning,
title={Learning to Extrapolate and Adjust: Two-Stage Meta-Learning for Concept Drift in Online Time Series Forecasting},
author={Zhaoyang Zhu and Weiqi Chen and YiFan Zhang and Qingsong Wen and Liang Sun},
year={2024},
url={https://openreview.net/forum?id=7U5QE9T4hI}
}

@article{von2020online,
  title={Online ensemble learning for load forecasting},
  author={Von Krannichfeldt, Leandro and Wang, Yi and Hug, Gabriela},
  journal={IEEE Transactions on Power Systems},
  volume={36},
  number={1},
  pages={545--548},
  year={2020},
  publisher={IEEE}
}

@article{hong2016probabilistic,
  title={Probabilistic electric load forecasting: A tutorial review},
  author={Hong, Tao and Fan, Shu},
  journal={International Journal of Forecasting},
  volume={32},
  number={3},
  pages={914--938},
  year={2016},
  publisher={Elsevier}
}

@article{ovadia2019can,
  title={Can you trust your model's uncertainty? evaluating predictive uncertainty under dataset shift},
  author={Ovadia, Yaniv and Fertig, Emily and Ren, Jie and Nado, Zachary and Sculley, David and Nowozin, Sebastian and Dillon, Joshua and Lakshminarayanan, Balaji and Snoek, Jasper},
  journal={Advances in neural information processing systems},
  volume={32},
  year={2019}
}

\end{document}